\documentclass[conference]{IEEEtran}
\IEEEoverridecommandlockouts
\usepackage{cite}
\usepackage{amsmath,amssymb,amsfonts}
\usepackage{graphicx}
\usepackage{textcomp}
\usepackage{balance} 
\usepackage{graphics} 
\usepackage{epsfig} 
\usepackage{times} 
\usepackage{subcaption}
\usepackage{newtxtext}
\usepackage{booktabs}  
\usepackage{multirow}  
\usepackage{array}     
\usepackage{float}
\usepackage{mathtools}
\usepackage{algorithm,algpseudocode}
\usepackage{amsthm}
\usepackage{newtxmath}
\newtheorem{definition}{Definition}

\usepackage{xcolor}
\def\BibTeX{{\rm B\kern-.05em{\sc i\kern-.025em b}\kern-.08em
    T\kern-.1667em\lower.7ex\hbox{E}\kern-.125emX}}
\begin{document}

\title{Joint Prediction of Human Motions and Actions in Human-Robot Collaboration \thanks{This paper has been submitted to IEEE International Conference on Advanced Intelligent Mechatronics (AIM).}
}

\author{\IEEEauthorblockN{Alessandra Bulanti}
\IEEEauthorblockA{
\textit{University of Genoa}, Italy \\
alessandra.bulanti@edu.unige.it}
\and
\IEEEauthorblockN{Alessandro Carfì}
\IEEEauthorblockA{
\textit{University of Genoa}, Italy \\
alessandro.carfi@unige.it}
\and
\IEEEauthorblockN{Fulvio Mastrogiovanni}
\IEEEauthorblockA{
\textit{University of Genoa}, Italy \\
fulvio.mastrogiovanni@unige.it}
}

\maketitle

\begin{abstract}
Fluent human--robot collaboration requires robots to continuously estimate human behaviour and anticipate future intentions.
This entails reasoning jointly about \emph{continuous movements} and \emph{discrete actions}, which are still largely modelled in isolation.
In this paper, we introduce \textsf{MA-HERP}, a hierarchical and recursive probabilistic framework for the \emph{joint estimation and prediction} of human movements and actions.
The model combines:
(i) a hierarchical representation in which movements compose into actions through admissible Allen interval relations,
(ii) a unified probabilistic factorisation coupling continuous dynamics, discrete labels, and durations, and
(iii) a recursive inference scheme inspired by Bayesian filtering, alternating top-down action prediction with bottom-up sensory evidence.
We present a preliminary experimental evaluation based on neural models trained on musculoskeletal simulations of reaching movements, showing accurate motion prediction, robust action inference under noise, and computational performance compatible with on-line human--robot collaboration.
\end{abstract}


\section{Introduction}
\label{sec:introduction}

\begin{figure}
\centering
\includegraphics[width=0.9\linewidth]{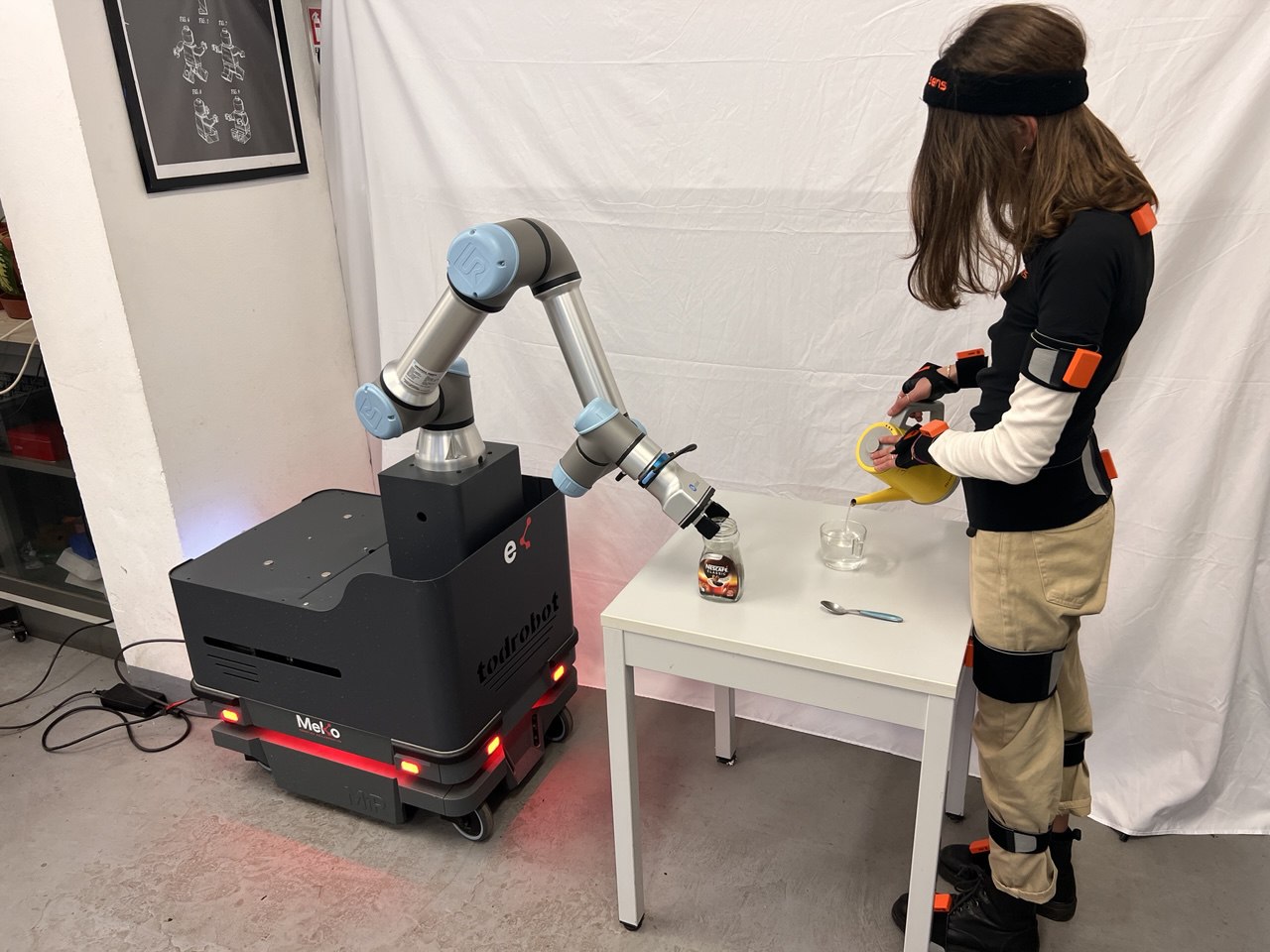}
\caption{
An illustrative example of a coffee-making scenario in which the prediction of human motions and actions can improve the perceived collaboration fluency to a great extent.
}
\label{fig:illustrative}
\end{figure}

Human--Robot Collaboration (HRC) requires robots to coordinate with humans under tight timing constraints, and the perceived \emph{fluency} of the interaction depends on how well the robot anticipates and adapts to human behaviour \cite{hoffman2019fluency}. 
To act proactively (for example, preparing a handover, synchronizing a shared manipulation, or enforcing safety margins before contact), a robot should continuously \emph{estimate} what the human is doing and \emph{predict} what will happen next (see Figure~\ref{fig:illustrative}). 
This demands reasoning over two intertwined descriptions of behaviour: 
(i) \emph{continuous movements}, that is, trajectories of body joints evolving in $\mathbb{R}^d$, and 
(ii) \emph{discrete actions}, that is, semantically meaningful events such as \texttt{reach}, \texttt{pick}, or \texttt{release}. 
Movements provide the geometric and dynamical detail needed for safe control and motion planning, while actions provide structure for task-level reasoning. 
However, these two levels have largely been treated in isolation \cite{Seminaraetal2023}.
A major difficulty is that HRC does not reduce to a single sequential stream of actions.
Humans often exhibit \emph{partial concurrency} (for example, grasping with one hand while repositioning another tool), and meaningful collaboration frequently depends on \emph{temporal semantics} between sub-events (for example, a grasp must occur \textsc{before} a handover; a lift may \textsc{meet} the end of a grasp). 
Moreover, composite actions typically emerge as structured compositions of simpler actions (for example, closing a door implies reaching the handle, grasping, rotating, and pulling, with admissible temporal relations). 
Capturing these phenomena requires models that can represent
(i) \emph{hierarchy} across abstraction levels, 
(ii) \emph{explicit temporal constraints} among intervals, and 
(iii) \emph{uncertainty} in both continuous trajectories and symbolic labels.

The literature offers strong results when movements or actions are considered separately.
Movement prediction is typically formulated as forecasting skeleton-based trajectories using recurrent, generative, or attention-based predictors \cite{ref_article35, ref_article39, ref_article40, ref_article47, ref_article55, ref_article53}, often improved by incorporating contextual cues such as object geometry, person--object relations, or affordances \cite{ref_article34, ref_article15, ref_article17}. 
Action prediction instead models sequences of discrete \textit{labels} using Markovian or recurrent architectures \cite{ref_article22, ref_article30, ref_article14}, again with context playing a key role through object detection, tracking, and attention mechanisms \cite{ref_article22, ref_article24, ref_article23}. 
Only few approaches attempt a \emph{joint} treatment of actions and movements, and they are commonly restricted to simple, mostly sequential tasks and limited action hierarchies \cite{ref_article18, ref_article19}. 
More general hybrid formalisms exist, for example, hierarchical temporal models such as Switching Linear Dynamical Systems and Hidden Semi-Markov Models \cite{ghahramani1997switching, yu2010hidden}, factor-graph formulations for structured probabilistic inference \cite{kschischang2001factor}, and symbolic temporal logics such as Allen's interval algebra \cite{Allen1983, vanBeek1992}, but they are rarely combined into an integrated, on-line estimation/prediction scheme tailored to HRC fluency.

In this paper, we address these gaps and introduce a unified framework for \textit{Movement-Action Hierarchical Estimation and Recursive Prediction}, which we refer to as \textsf{MA-HERP}, for the \emph{joint estimation and prediction} of movements and actions in HRC scenarios. 
\textsf{MA-HERP} is grounded on three design principles:
(i) \textit{hierarchical representation}:
continuous movements correspond to basic action intervals, which compose into higher-level composite actions through admissible temporal relations, enabling partial orders and controlled overlaps;
(ii) \textit{recursive estimation and prediction}: an on-line prediction--likelihood--update cycle propagates beliefs over time, alternating top-down discrete forecasts with bottom-up continuous evidence, while enforcing temporal and duration constraints;
(iii) \textit{joint probabilistic coupling}: movements and actions are modelled within a single factorized distribution, allowing sensory data to update beliefs about actions and, likewise, action beliefs to condition motion interpretation and prediction.

\textsf{MA-HERP} is not intended as a novel standalone movement predictor or action classifier. 
Rather, it is a joint hierarchical continuous--discrete modelling framework that integrates Allen-relation-based temporal constraints, multi-level action compositions, and a unified probabilistic factorization.
Our contributions are: 
(i) a formal hierarchical representation relating movements to actions via interval-based temporal relations with explicit durations; 
(ii) a joint probabilistic model coupling continuous movements and discrete symbols in factor-graph form; 
(iii) a recursive estimation and prediction algorithm approximating the joint model.
Section~\ref{sec:representation} introduces the hierarchical representation,
Section~\ref{sec:joint} details the joint model and recursive estimation/prediction scheme, whereas
a preliminary evaluation using musculoskeletal simulations and data-driven components is discussed in Section~\ref{sec:evaluation}.
Conclusions follow.

\section{Hierarchical Representation of Movements and Actions}
\label{sec:representation}

We introduce the notation used throughout the paper.
Let $t \in \mathbb{N}$ be a discrete time index with sampling period $\Delta t$.
We denote by $X_t \in \mathbb{R}^d$ the continuous observation vector at time $t$, and by
$X_{1:T} \coloneqq (X_1, \dots, X_T)$ a sequence over a temporal window.
In \textsf{MA-HERP}, $X_t$ may include joint positions, velocities, and accelerations of body parts.
We first define an \emph{action interval}.

\begin{definition}
An elementary (or composite) action $a_k$ is represented by a labelled interval
\begin{equation}
a_k \;=\; (\ell_k, s_k, e_k),    
\label{eq:action}
\end{equation}
where $\ell_k \in \mathcal{L}$ is a discrete label, $s_k,e_k \in \{1,\dots,T\}$ are start/end instants with
$1 \le s_k < e_k \le T$, and $d_k = e_k - s_k + 1$ is the action duration.
\end{definition}

\noindent
Temporal relations between action intervals are modeled through Allen's interval algebra \cite{Allen1983, vanBeek1992}.

\begin{definition}
Given two intervals $a_i=(\ell_i,s_i,e_i)$ and $a_j=(\ell_j,s_j,e_j)$, their temporal relation
$\mathcal{R}(a_i,a_j)$ belongs to the canonical Allen set
\begin{equation}
\begin{split}
\mathfrak{A} = \{&\textsc{before}, \textsc{meets}, \textsc{overlaps}, 
\textsc{starts}, \textsc{during},\\
&\textsc{finishes}, \textsc{equals}\},
\end{split}
\end{equation}
together with the corresponding converses.
For instance, $\textsc{before}$ is induced by $e_i < s_j$, and $\textsc{meets}$ by $e_i = s_j$.
\end{definition}

\noindent
In real motions, not all Allen relations are physically plausible (for example, two distinct same-hand actions cannot overlap).
We capture feasible temporal patterns via \emph{configurations}.

\begin{definition}
A configuration $A$ denotes a finite set of action intervals constituting a composite action:
\begin{equation}
A = \{a_1, a_2, \dots, a_K\}.
\label{eq:configuration}
\end{equation}
For any pair $(i,j)$, the relation $\mathcal{R}(a_i,a_j)$ is induced by the corresponding boundaries $(s_i,e_i)$ and $(s_j,e_j)$.
\end{definition}

\noindent
Admissible temporal patterns restrict Allen relations.

\begin{definition}
Given an admissible set $\mathfrak{A}_{\text{A}} \subseteq \mathfrak{A}$, a configuration $A$ is \emph{admissible} if, for every pair $(i,j)$, it holds that
$\mathcal{R}(a_i,a_j) \in \mathfrak{A}_{\text{A}}(\ell_i,\ell_j)$.
\end{definition}

\noindent
The admissible set may depend on the labels and their physical semantics, that is,
$\mathfrak{A}_{\text{A}}(\ell_i,\ell_j)\subseteq\mathfrak{A}$.
For instance, overlaps can be disallowed for same-hand actions while being permitted across different hands.
We now introduce the hierarchy that integrates movements and actions.

\begin{figure}[t]
\centering
\includegraphics[width=0.8\linewidth]{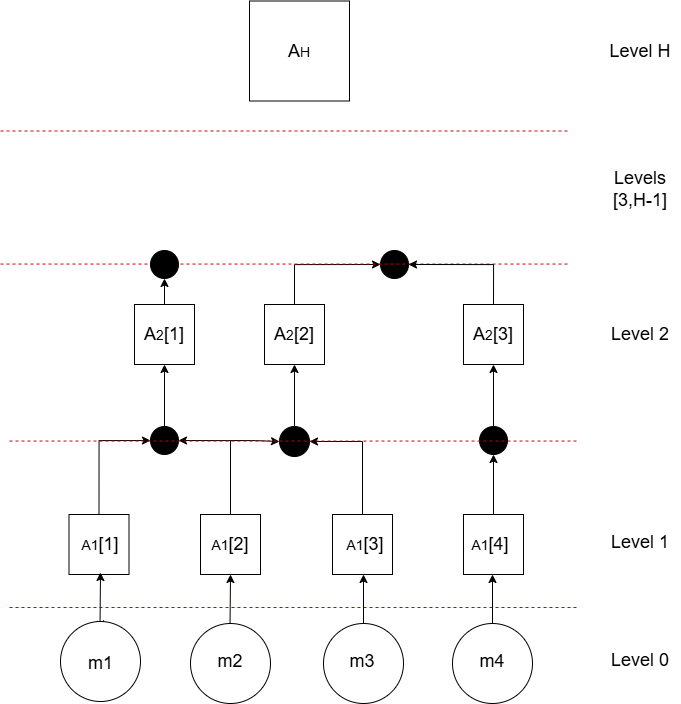}
\caption{
Hierarchical structure of movements and actions \textsf{MA-HERP}.
Level $h=0$ encodes continuous movement segments, while levels $h\ge1$ encode discrete actions. 
Edges denote temporal relations defined via Allen's interval algebra.}
\label{fig:Hierarchy}
\end{figure}

In \textsf{MA-HERP}, we consider $H{+}1$ hierarchical levels, $h=0,\dots,H$.
Level $h=0$ contains \emph{movement segments}, that is, continuous trajectories in $\mathbb{R}^d$ \cite{Lastricoetal2024}.
Levels with $h\ge1$ aggregate lower-level intervals via Allen relations to form increasingly abstract \emph{actions}.
An illustration is provided in Figure~\ref{fig:Hierarchy}.

\begin{definition}
For each level $h=0,\dots,H$, we define a finite set of intervals
\begin{equation}
A^{h} = \{a^{h}_1, a^{h}_2, \dots, a^{h}_{K_h}\},
\label{eq:hierarchical}
\end{equation}
where each interval $a^{h}_k$ is specified by:
(i) a label $\ell^{h}_k \in \mathcal{L}^{h}$,
(ii) a time range $(s^{h}_k,e^{h}_k)$, and
(iii) a (possibly empty) child set $\mathcal{C}(a^{h}_k) \subseteq A^{h-1}$ such that, for any child pair $c_1,c_2 \in \mathcal{C}(a^{h}_k)$, the Allen relation $\mathcal{R}(c_1,c_2)$ is admissible according to a level-specific composition table $\mathbf{M}^{h}$.
\noindent
Formally, $\mathbf{M}^{h}$ maps pairs of child labels to admissible relations, that is, 
$\mathbf{M}^{h}:\mathcal{L}^{h-1}\times\mathcal{L}^{h-1}\rightarrow 2^{\mathfrak{A}}$.
\end{definition}

\noindent
A higher-level (composite) action is thus a temporal pattern of lower-level intervals consistent with $\mathbf{M}^h$.
At level $h=0$, intervals are aligned with continuous observations, that is,
$a^{0}_k \leftrightarrow X_{s^{0}_k:e^{0}_k}$.

As an example, consider a composite action at level $h=2$ with label \texttt{handover}.
Its child set $\mathcal{C}(a^{2}_k)$ may include level-1 intervals such as
\texttt{reach(object)}, \texttt{grasp(object)}, \texttt{reach(handover\_pose)}, and \texttt{release(object)}.
A valid pattern (declared by $\mathbf{M}^2$) requires:
(i) \texttt{reach(object)} \textsc{before} \texttt{grasp(object)}, then
(ii) \texttt{grasp(object)} \textsc{before} \texttt{reach(handover\_pose)}, and finally 
(iii) \texttt{reach(handover\_pose)} \textsc{meets} \texttt{release(object)}.
In general, the child set specifies \emph{which} lower-level components are used, while $\mathbf{M}^h$ specifies \emph{how} they are temporally arranged.

To encode admissibility in the probabilistic model of Section~\ref{sec:joint}, we translate Boolean constraints into numerical factors.

\begin{definition}
For each pair of action intervals $(a_i,a_j)$ we define a plausibility factor
\begin{equation}
\psi_{\mathcal{R}}(a_i,a_j)
\;=\;
\begin{cases}
1, & \text{if } \mathcal{R}(a_i,a_j)\in \mathfrak{A}_{\text{A}}(\ell_i,\ell_j), \\
0, & \text{otherwise.}
\end{cases}
\label{eq:plausibility}
\end{equation}
\end{definition}

\noindent
A soft/fuzzy version can be adopted to tolerate segmentation noise \cite{BuoncompagniMastrogiovanni2024}.
For instance, if (\texttt{reach},\texttt{grasp}) should be \textsc{before}, then
$\psi_{\textsc{before}}(\texttt{reach},\texttt{grasp})=1$ (hard case), or
$\psi_{\textsc{before}}(\texttt{reach},\texttt{grasp})\approx 0.9$ (soft case).

Having defined discrete intervals and their temporal relations, we relate them to continuous observations.
We assume that the continuous dynamics regime depends on the instantaneous discrete label:
\begin{equation}
p(X_t \mid X_{t-1}, \ell_t, C_t; \theta).
\label{eq:dynamics}
\end{equation}
Here, $C_t$ denotes optional contextual information (for example, objects, affordances, or a scene graph) available at time $t$, and $\theta$ are the parameters of the chosen dynamics model.
The mapping from time to labels is induced by the intervals: if $t \in [s_k,e_k]$ for some $a_k=(\ell_k,s_k,e_k)$, then $\ell_t=\ell_k$.
In the case of parallel action streams (for example left/right hands), $\ell_t$ becomes a tuple/vector of labels.

On the one hand, labels (and intervals) condition the prediction of continuous observations.
On the other hand, observations provide evidence constraining the inference of intervals and their temporal relations.
This bidirectional coupling is the core idea of the joint probabilistic model described in the next Section.

\section{Recursive Estimation and Prediction on a Joint Movement-Action Model}
\label{sec:joint}

The goal of \textsf{MA-HERP} is the \emph{joint estimation} and \emph{prediction} of movements and actions under a single hybrid model.
We assume time-aligned and normalized observations $X_t$, context streams $C_t$, parametrized label transitions $p(\ell\mid \ell',C)$ and duration models $p(d\mid \ell)$, with hazards $h_\ell$ to avoid unrealistic memoryless segmentations.
Temporal compatibility is enforced through label-dependent admissible sets $\mathfrak{A}_{\text{A}}(\ell_i,\ell_j)$, applied selectively to relevant pairs to keep inference tractable.

Let $A^{(0:H)} \coloneqq \{A^{h}\}_{h=0}^H$ denote the collection of interval sets across levels.
\textsf{MA-HERP} operates in two modes.
\emph{Recognition} estimates the most plausible past segmentation:
\begin{equation}
\widehat{A}^{(0:H)} 
= \arg\max \; p\!\left(A^{(0:H)} \mid X_{1:T}, C_{1:T}\right).
\label{eq:estimates}
\end{equation}
\emph{Prediction} forecasts future motions (and, optionally, future intervals) under the same joint model:
\begin{equation}
\widehat{X}_{T+1:T+\Delta}
= \arg\max \; p\!\left(X_{T+1:T+\Delta} \mid \widehat{A}^{(0:H)}, X_{1:T}, C_{1:T}\right).
\label{eq:predicts}
\end{equation}

We express the joint model via a factor graph \cite{kschischang2001factor}:
\begin{align}
p\!\left(A^{(0:H)}, X_{1:T} \mid C_{1:T}\right) \propto 
& \prod_{h=0}^H \prod_{(i,j)\in \mathcal{P}^{(h)}} 
\psi_{\mathcal{R}}\!\left(a^{h}_i, a^{h}_j\right) \nonumber\\
& \times \prod_{h=1}^H \prod_{k} 
\psi_{\text{C}}\!\left(a^{h}_k, \mathcal{C}(a^{h}_k)\right) \nonumber\\
& \times \prod_{t=1}^T p\!\left(X_t \mid X_{t-1}, \ell_t, C_t; \theta\right) \nonumber\\
& \times \prod_{k} p\!\left(\ell_k, d_k \mid \ell_{k-1}, C_{s_k:e_k}; \eta\right),
\label{eq:joint}
\end{align}
where the factors encode, respectively: (i) Allen-based plausibility between interval pairs (Equation~\ref{eq:plausibility}); (ii) hierarchical composition consistency; (iii) label-conditioned continuous dynamics; and (iv) semi-Markov label transitions with explicit durations \cite{yu2010hidden}.
Prediction (Equation~\ref{eq:predicts}) inherits the same factorization, restricted to future variables and conditioned on $X_{1:T},C_{1:T}$.

Exact inference in Equation~\ref{eq:joint} is intractable due to the combinatorial number of configurations across levels and temporal relations.
We therefore adopt an on-line recursive approximation, summarized as a mapping
\begin{equation}
\mathcal{F}_\Theta:\; (\Pi_t, X_{1:t}, C_{1:t}) \mapsto
(\widehat{\Pi}_{t+1:t+\Delta}, \widehat{X}_{t+1:t+\Delta}, S_{t+1}),
\label{eq:recursive}
\end{equation}
where $\Pi_t$ is an inference summary (for example, beliefs over active labels/levels and boundary/duration statistics), $S_{t+1}$ is an internal state, and $\Theta$ collects all model parameters.

\noindent
\textit{Single-chain recursion.}
Let $\pi_{t|t-1}(\ell)$ be the predictive prior over labels, $\lambda_t(\ell)$ the observation likelihood, and $\alpha_t(\ell)$ the posterior belief that the instantaneous label equals $\ell$ at time $t$.
A compact prediction--likelihood--update cycle reads:
\begin{align}
\pi_{t|t-1}(\ell) 
&\propto \sum_{\ell'} p\!\left(\ell \mid \ell', C_{t-1}; \eta\right)\,\alpha_{t-1}(\ell'), 
\label{eq:pred}\\
\lambda_t(\ell) 
&\propto p\!\left(X_t \mid X_{t-1}, \ell, C_t; \theta\right), 
\label{eq:like}\\
\alpha_t(\ell) 
&\propto \pi_{t|t-1}(\ell)\,\lambda_t(\ell)\,
\Phi_{\text{A},t}(\ell;\Pi_{t-1}),
\label{eq:update}
\end{align}
where $\Phi_{\text{A},t}$ penalizes label assignments inconsistent with active Allen constraints (and, in practice, with boundary/duration bookkeeping).
Durations are handled via the HSMM hazard $h_\ell(\cdot)$, modulating the probability of closing/opening intervals.

\noindent
\textit{Multi-chain / multi-level case.}
With parallel chains and multiple levels, $\ell$ becomes a tuple, and $\Phi_{\text{A},t}$ factorizes over selected intra-/inter-chain pairs, enforcing only locally relevant constraints.

\begin{algorithm}[t!]
\caption{\textsf{MA-HERP}}
\label{alg:hrf}
\begin{algorithmic}[1]
\Require $\Theta=(\theta,\eta)$; admissible sets $\mathfrak{A}_{\text{A}}$; duration priors $p(d\mid\ell)$; $S_1,\Pi_1$
\For{$t=1$ \textbf{to} $T$}
\State $\pi_{t|t-1}(\ell) \propto \sum_{\ell'} p(\ell\mid \ell',C_{t-1};\eta)\,\times\,\alpha_{t-1}(\ell')$ \Comment{Eq.\ref{eq:pred}}
\State $\lambda_t(\ell) \propto p(X_t\mid X_{t-1},\ell,C_{t};\theta)$ \Comment{Eq.\ref{eq:like}}
\State $\alpha_t(\ell) \propto \pi_{t|t-1}(\ell)\,\times\,\lambda_t(\ell)\,\times\,\Phi_{\text{A},t}(\ell;\Pi_{t-1})$ \Comment{Eq.\ref{eq:update}}
\State $(\hat{s},\hat{e}) \leftarrow \textsc{DurationUpdate}(h_\ell)$
\State $\widehat{X}_{t+1} \leftarrow \mathbb{E}[X_{t+1}\mid X_t,\ell^\star_t,C_{t+1};\theta]$
\State $\Pi_{t+1} \leftarrow \textsc{InferenceUpdate}(\Pi_t,\alpha_t,\hat{s},\hat{e})$
\State $S_{t+1}\leftarrow\textsc{StateUpdate}(S_t)$
\EndFor
\end{algorithmic}
\end{algorithm}

Algorithm~\ref{alg:hrf} summarizes the on-line recursion: it propagates label beliefs, scores observations under label-conditioned dynamics, applies Allen consistency, updates boundaries/durations, and produces the next-step trajectory prediction.

\section{Experimental Evaluation}
\label{sec:evaluation}

This Section presents an initial evaluation of \textsf{MA-HERP}, designed as a \emph{foundational} validation of the modelling assumptions introduced in Sections~\ref{sec:representation}--\ref{sec:joint}.
While the experimental setup is intentionally simplified, it targets quantities that are directly relevant to HRC scenarios: 
(i) the ability to anticipate \emph{human motion} early enough to support proactive safety and timing, and 
(ii) the ability to infer \emph{action-level intent} consistently over time, so that a robot can coordinate its assistance under \textit{fluency} constraints \cite{hoffman2019fluency}.
In particular, we quantify: 
(i) \emph{movement prediction} accuracy for the continuous dynamics term in Equation~\eqref{eq:dynamics}, and 
(ii) \emph{action prediction and classification} performance for the discrete component of Equation~\eqref{eq:joint}, with special attention to robustness under sensory perturbations.
Finally, we provide a short end-to-end example showing how the recursive loop of Algorithm~\ref{alg:hrf} behaves over a sequence of reaching movements, as a proxy for basic pick-and-place routines typical of HRC scenarios.

\begin{figure}[t!]
\centering
\begin{minipage}[b]{0.15\textwidth}
\centering
\includegraphics[width=\linewidth]{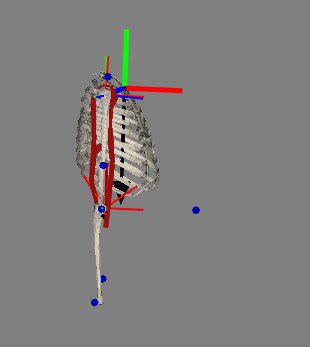}
\end{minipage}
\begin{minipage}[b]{0.15\textwidth}
\centering
\includegraphics[width=\linewidth]{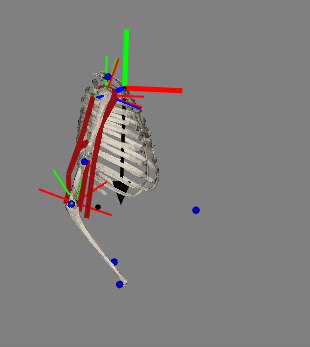}
\end{minipage}
\begin{minipage}[b]{0.15\textwidth}
\centering
\includegraphics[width=\linewidth]{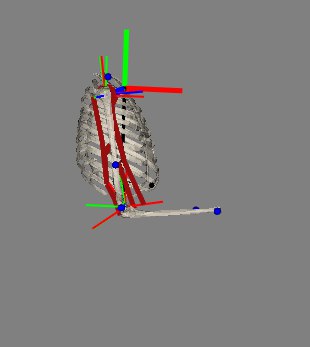}
\end{minipage}
\caption{
A musculoskeletal model executing a reaching movement. 
Three poses are shown from left to right.}
\label{fig:skeletonsequence}
\end{figure}

\noindent
\textit{Setup and data generation}.
To generate controlled motion datasets, we employ a musculoskeletal model of a human torso and right arm from \textit{Bioptim} \cite{Michaudetal2021}, which features two degrees of freedom, six muscles, and an optimal-control-based controller.
We introduce an additional degree of freedom at the elbow joint, enabling motion in a three-dimensional workspace. 
We focus on reaching movements, and Bioptim is used to compute an optimal trajectory that reaches a desired target in configuration space.
Trajectories are generated by minimising a cost function composed of a Lagrange term integrated over time and a Mayer term defined on the trajectory as a whole, which penalises the Euclidean distance between a virtual \textit{marker} on the wrist and the target marker (blue in Figure~\ref{fig:skeletonsequence}).
This setting provides a clean testbed to evaluate whether the proposed \emph{joint continuous--discrete} machinery can:
(i) preserve stable motion forecasts, and 
(ii) produce action beliefs that converge early enough to be useful for coordination.

We consider three distinct reaching motions defined by three target volumes \textsf{A}, \textsf{B}, and \textsf{C} (each a $5$ cm side cube) within the reachable workspace, and an initial pose \textsf{I}. 
The centre of each target volume is positioned (in metres) as:
$T_\textsf{A} = (0.20, \text{--}0.30, 0.15)$, 
$T_\textsf{B} = (0.40, \text{--}0.10, 0.20)$, and 
$T_\textsf{C} = (0.15, 0.05, 0.10)$.
Within each volume, $1000$ target points are sampled on a uniform 3D grid at $5$ mm resolution (that is, $1000$ points per target cube).
This yields a labelled set of reaching trajectories that can be interpreted as low-level motion primitives.
We define \emph{nine} labels $\ell_k$ by considering reaches from the initial pose \textsf{I} and all pairwise transitions between target volumes.
Specifically: 
(i) the first three labels capture movements starting from \textsf{I} and ending in each target cube,
$\ell_1 = \textsf{I}\to\textsf{A}$,
$\ell_2 = \textsf{I}\to\textsf{B}$, and
$\ell_3 = \textsf{I}\to\textsf{C}$; 
(ii) the remaining six labels represent pairwise reaches between target volumes, using the final poses of (i) as initial conditions:
$\ell_4 = \textsf{A}\to\textsf{B}$,
$\ell_5 = \textsf{A}\to\textsf{C}$,
$\ell_6 = \textsf{B}\to\textsf{A}$,
$\ell_7 = \textsf{B}\to\textsf{C}$,
$\ell_8 = \textsf{C}\to\textsf{A}$, and
$\ell_9 = \textsf{C}\to\textsf{B}$.
We consider three joint angles, $q_1,q_2,q_3$, hence $X_t \in \mathbb{R}^3$ and the predictors operate in a three-dimensional configuration space.
For each label, we generate $1000$ trajectories, each with $3 \times 251$ samples (thus $T=251$ is fixed and does not need to be estimated by \textsf{MA-HERP} in this experiment).
From the original noise-free dataset, we derive additional datasets by injecting uniform additive noise along the trajectory at 10\%, 30\%, and 50\%.
Here we focus on the noise-free dataset \textsf{D-0} and on a combined dataset \textsf{D-0-10-30} pooling trajectories from 0\%, 10\%, and 30\% noise levels.
At level $h{=}0$, each action interval $a^0_k=(\ell_k,1,251)$ indexes the label-conditioned continuous dynamics regime for $X_{1:251}$.
At $h \ge 1$, admissible temporal relations (used later in the end-to-end example) are constrained by $\psi_{\mathcal{R}}$ and by the corresponding composition table $\mathbf{M}^{h}$.

\begin{figure}[t!]
\centering
\begin{subfigure}{0.23\textwidth}
\centering
\includegraphics[width=\linewidth]{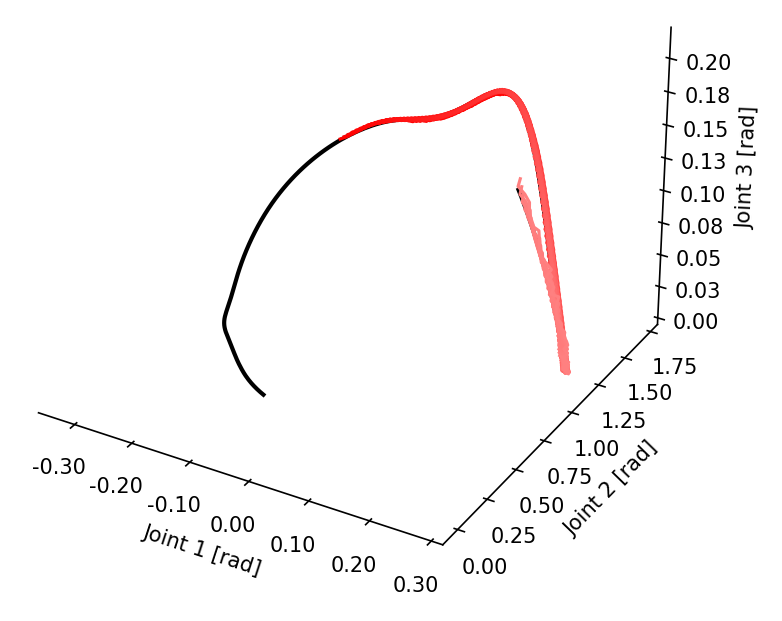}
\caption{}
\label{fig:prediction}
\end{subfigure}
\begin{subfigure}{0.23\textwidth}
\centering
\includegraphics[width=\linewidth]{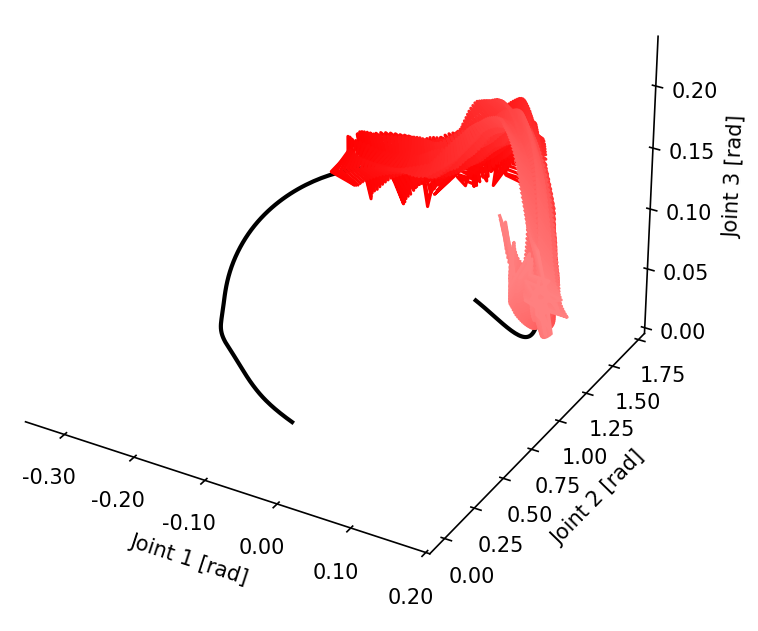}
\caption{}
\label{fig:prediction30}
\end{subfigure}
\caption{
Predictions (red) in configuration space for $\ell_1 = \textsf{I}\to\textsf{A}$ obtained with (\subref{fig:prediction}) the model trained on \textsf{D-0} and (\subref{fig:prediction30}) the model trained on \textsf{D-0-10-30}, compared against ground truth (black).}
\label{fig:predictionsgraphs}
\end{figure}

\noindent
\textit{Movement prediction}.
To implement line~3 in Algorithm~\ref{alg:hrf}, we train two label-conditioned autoregressive Transformer models, respectively on \textsf{D-0} and \textsf{D-0-10-30}.
For \textsf{D-0}, we use $900$ trajectories per label for training and $100$ for testing; for \textsf{D-0-10-30}, we use $2700$ for training and $300$ for testing.
We consider temporal windows of $W=100$ steps to forecast the next $P=50$ steps.
At inference time, a sliding window with stride $1$ produces predictions for every $t\ge 100$.
Qualitative results are shown in Figure~\ref{fig:predictionsgraphs}.
When trained on \textsf{D-0}, predicted trajectories are visually indistinguishable from the ground truth.
When trained on \textsf{D-0-10-30}, the injected noise yields more scattered forecasts, which is expected and highlights the importance of uncertainty management in HRC.

\begin{table*}[t!]
\centering
\caption{
Pearson Correlation Coefficient (PCC) and Root Mean Square Error (RMSE) for all label-conditioned movement predictors over all joint angles ($q_1,q_2,q_3$), for \textsf{D-0} and \textsf{D-0-10-30}. Reported values are rounded to two decimal places.
}
\label{tab:table1}
\begin{tabular}{c|*{4}{c}}
\multicolumn{5}{c}{\textsf{D-0}} \\ 
\hline
\hline
\textbf{Models} & \textbf{PCC (Avg.)} [-1,1] & \textbf{PCC (Std.)} [0,1] & \textbf{RMSE (Avg.)} [rad] & \textbf{RMSE (Std.)} [rad] \\ 
\hline
$\textsf{I}\to\textsf{A}$ & (0.99, 0.99, 0.99)  & (0.01, 0.02, 0.06)  & (0.01, 0.01, 0.01)  & (0.01, 0.01, 0.01)  \\
$\textsf{I}\to\textsf{B}$ & (0.98, 0.86, 0.21)  & (0.06, 0.32, 0.78)  & (0.02, 0.01, 0.01)  & (0.01, 0.01, 0.01)  \\
$\textsf{I}\to\textsf{C}$ & (0.99, 0.99, 0.99)  & (0.01, 0.01, 0.33) & (0.01, 0.01, 0.01) & (0.01, 0.01, 0.01) \\
$\textsf{A}\to\textsf{B}$ & (0.99, 0.98, 0.62) & (0.01, 0.05, 0.60) & (0.01, 0.01, 0.01) & (0.01, 0.01, 0.01) \\
$\textsf{A}\to\textsf{C}$ & (0.99, 0.99, 0.99) & (0.01, 0.01, 0.02) & (0.01, 0.02, 0.01) & (0.01, 0.01, 0.01) \\
$\textsf{B}\to\textsf{A}$ & (0.99, 0.94, 0.96) & (0.02, 0.17, 0.14) & (0.01, 0.02, 0.01) & (0.01, 0.02, 0.01) \\
$\textsf{B}\to\textsf{C}$ & (0.85, 0.99, 0.95) & (0.24, 0.04, 0.12) & (0.02, 0.04, 0.01) & (0.01, 0.02, 0.01) \\
$\textsf{C}\to\textsf{A}$ & (0.99, 0.96, 0.89) & (0.04, 0.08, 0.27) & (0.01, 0.02, 0.01) & (0.01, 0.02, 0.01) \\
$\textsf{C}\to\textsf{B}$ & (0.99, 0.93, 0.69) & (0.03, 0.11, 0.41) & (0.01, 0.02, 0.01) & (0.01, 0.02, 0.01) \\ 
\hline
\multicolumn{5}{c}{\textsf{D-0-10-30}} \\ 
\hline
\hline
\textbf{Models} & \textbf{PCC (Avg.)} [-1,1] & \textbf{PCC (Std.)} [0,1] & \textbf{RMSE (Avg.)} [rad] & \textbf{RMSE (Std.)} [rad] \\ 
\hline
$\textsf{I}\to\textsf{A}$ & (0.37, 0.48, 0.36) & (0.40, 0.40, 0.21) & (0.25, 0.29, 0.25) & (0.21, 0.25, 0.21) \\
$\textsf{I}\to\textsf{B}$ & (0.51, 0.15, 0.14) & (0.37, 0.32, 0.28) & (0.28, 0.28, 0.23) & (0.25, 0.25, 0.20) \\
$\textsf{I}\to\textsf{C}$ & (0.55, 0.79, 0.35) & (0.38, 0.24, 0.39) & (0.20, 0.29, 0.24) & (0.17, 0.24, 0.20) \\
$\textsf{A}\to\textsf{B}$ & (0.55, 0.28, 0.23) & (0.37, 0.37, 0.35) & (0.27, 0.28, 0.24) & (0.24, 0.25, 0.20) \\
$\textsf{A}\to\textsf{C}$ & (0.56, 0.80, 0.38) & (0.39, 0.23, 0.40) & (0.20, 0.29, 0.25) & (0.17, 0.24, 0.20) \\
$\textsf{B}\to\textsf{A}$ & (0.43, 0.42, 0.36) & (0.41, 0.41, 0.42) & (0.25, 0.29, 0.25) & (0.21, 0.25, 0.21) \\
$\textsf{B}\to\textsf{C}$ & (0.44, 0.73, 0.32) & (0.41, 0.29, 0.37) & (0.20, 0.31, 0.25) & (0.17, 0.24, 0.20) \\
$\textsf{C}\to\textsf{A}$ & (0.42, 0.46, 0.15) & (0.39, 0.38, 0.32) & (0.26, 0.30, 0.24) & (0.22, 0.25, 0.20) \\
$\textsf{C}\to\textsf{B}$ & (0.62, 0.31, 0.14) & (0.33, 0.37, 0.30) & (0.27, 0.29, 0.24) & (0.23, 0.25, 0.20) \\ \hline
\end{tabular}
\end{table*}

\noindent
\textit{Evaluation metrics and results}. To quantify prediction sensitivity to noise, we compute the Pearson Correlation Coefficient (PCC) and the Root Mean Square Error (RMSE) between predicted and ground-truth trajectories, for all label-conditioned models and for all joints (Table~\ref{tab:table1}).
As expected, the model trained on \textsf{D-0} achieves near-perfect PCC on most trajectories, with occasional drops associated with highly non-linear segments (for example, abrupt changes in direction).
RMSE values are consistently low, with negligible variance.
When training on \textsf{D-0-10-30}, PCC decreases and variance increases, and RMSE correspondingly grows, confirming that measurement perturbations directly affect the stability of motion forecasting.
From an HRC perspective, this supports a key requirement, namely when sensory conditions deteriorate, predictive modules should expose increased uncertainty (or degraded accuracy) so that the robot can react by adopting safer margins and more conservative timing.

\begin{table}[t]
\centering
\caption{
Average and standard deviation inference times for movement prediction (forecasting $P=50$ steps) for all label-conditioned models, for \textsf{D-0} (100 test trajectories) and \textsf{D-0-10-30} (300 test trajectories).}
\label{tab:group_stats_datasets}
\begin{tabular}{ccc|ccc}
\multicolumn{3}{c}{\textsf{D-0}} & \multicolumn{3}{c}{\textsf{D-0-10-30}} \\
\hline
\hline
\textbf{Model} & \textbf{Avg.} [s] & \textbf{Std.} [s] & \textbf{Model} & \textbf{Avg.} [s] & \textbf{Std.} [s] \\ 
\hline
$\textsf{I}\to\textsf{A}$ & 0.1363 & 0.0130 & $\textsf{I}\to\textsf{A}$ & 0.1753 & 0.0097 \\
$\textsf{I}\to\textsf{B}$ & 0.1457 & 0.0013 & $\textsf{I}\to\textsf{B}$ & 0.1783 & 0.0027 \\
$\textsf{I}\to\textsf{C}$ & 0.1448 & 0.0017 & $\textsf{I}\to\textsf{C}$ & 0.1777 & 0.0019 \\
$\textsf{A}\to\textsf{B}$ & 0.1359 & 0.0131 & $\textsf{A}\to\textsf{B}$ & 0.1452 & 0.0080 \\
$\textsf{A}\to\textsf{C}$ & 0.1382 & 0.0133 & $\textsf{A}\to\textsf{C}$ & 0.1433 & 0.0082 \\
$\textsf{B}\to\textsf{A}$ & 0.1462 & 0.0019 & $\textsf{B}\to\textsf{A}$ & 0.1462 & 0.0025 \\
$\textsf{B}\to\textsf{C}$ & 0.1385 & 0.0132 & $\textsf{B}\to\textsf{C}$ & 0.1427 & 0.0089 \\
$\textsf{C}\to\textsf{A}$ & 0.1473 & 0.0034 & $\textsf{C}\to\textsf{A}$ & 0.1460 & 0.0018 \\
$\textsf{C}\to\textsf{B}$ & 0.1384 & 0.0125 & $\textsf{C}\to\textsf{B}$ & 0.1420 & 0.0084 \\ 
\hline
\end{tabular}
\end{table}
\noindent
\textit{Computational cost}. 
We measure the time required to predict the next $P=50$ steps for all label-conditioned movement predictors.
Results, obtained on an Intel(R) Core(TM) i7-10700 CPU @ 2.90 GHz, 32 GB RAM, are reported in Table~\ref{tab:group_stats_datasets}.
Average inference times are well below one second, enabling proactive behaviours such as collision avoidance and timely assistance rather than purely reactive control.
Standard deviations are small and relatively homogeneous across labels, indicating stable computational behaviour across different motion regimes.

\noindent
\textit{Action prediction and classification}.
We evaluate the discrete component of \textsf{MA-HERP}, that is, the ability to infer a semantically meaningful label from an evolving motion stream.
In HRC terms, this corresponds to estimating \emph{intent} early enough to support proactive coordination rather than classifying only once the motion is completed.
In our simplified testbed, each trajectory is associated with a composite action label indicating the target volume, namely \texttt{reach-a}, \texttt{reach-b}, and \texttt{reach-c}.
Accordingly, we train four Transformer-based classifiers reflecting different initial conditions:
(i) $\textsf{I}\to\textsf{ABC}$ classifies all trajectories starting at pose \textsf{I} and reaching any volume,
(ii) $\textsf{A}\to\textsf{BC}$ classifies trajectories leaving volume \textsf{A} towards either \textsf{B} or \textsf{C},
(iii) $\textsf{B}\to\textsf{AC}$ classifies trajectories leaving \textsf{B}, and
(iv) $\textsf{C}\to\textsf{AB}$ classifies trajectories leaving \textsf{C}.
Taken together, these classifiers instantiate the last factor in Equation~\eqref{eq:joint} (where durations $d_k$ are known in advance in this experiment), and provide the belief $\alpha_t(\ell)$ that can be used within the recursive loop.
To analyse robustness under sensory perturbations, for each classifier we train two versions, one on \textsf{D-0} and one on \textsf{D-0-10-30}.
For \textsf{D-0}, model $\textsf{I}\to\textsf{ABC}$ uses $2700$ trajectories for training and $300$ for testing, while each of the other three models uses $1800$ trajectories for training and $200$ for testing.
For \textsf{D-0-10-30}, the corresponding training and test sets are three times larger, as the dataset pools the 0\%, 10\%, and 30\% noise regimes.

\noindent
\textit{Incremental vs.\ sliding-window inference}.
At inference time we consider two conditions that emulate on-line acquisition, differing in how much history is retained.
In the first, which we refer to as \textit{incremental prefix}, we perform incremental classification by providing each model with increasingly larger prefixes of the trajectory, using $10$-sample increments, for example, $X_{1:10}, X_{1:20}, X_{1:30}, \dots$ until completion.
This condition reflects an optimistic setting in which \textsf{MA-HERP} can maintain and exploit an expanding memory of past observations, and it directly measures \emph{how early} action evidence becomes discriminative.
In the second, called \textit{sliding window}, we use a fixed window of $10$ samples, shifted by one time step, that is, $X_{1:10}, X_{2:11}, X_{3:12}, \dots$.
The final label is obtained by majority vote over the sequence of window-wise predictions.
This condition is closer to real-time pipelines where computation and buffering constraints impose limited memory; it also stresses the sensitivity of the classifiers to local trajectory features.

\begin{table}[t!]
\centering
\caption{Classification accuracy (majority vote for sliding-window inference).}
\label{tab:accuracy}
\begin{tabular}{cc|cc}
\multicolumn{2}{c}{\textsf{D-0}} & \multicolumn{2}{c}{\textsf{D-0-10-30}} \\
\hline
\hline
\textbf{Model} & \textbf{Accuracy} [0,1] & \textbf{Model} & \textbf{Accuracy} [0,1] \\ 
\hline
$\textsf{I}\to\textsf{A}\textsf{B}\textsf{C}$ & 0.95 & $\textsf{I}\to\textsf{A}\textsf{B}\textsf{C}$ & 0.66 \\
$\textsf{A}\to\textsf{BC}$ & 1.00 & $\textsf{A}\to\textsf{BC}$ & 1.00 \\
$\textsf{B}\to\textsf{AC}$ & 0.89 & $\textsf{B}\to\textsf{AC}$ & 0.50 \\
$\textsf{C}\to\textsf{AB}$ & 0.61 & $\textsf{C}\to\textsf{AB}$ & 0.80 \\ 
\hline
\end{tabular}
\end{table}

\begin{figure*}[t]
\centering
\caption{Confusion matrices for \textsf{D-0} (top row) and \textsf{D-0-10-30} (bottom row) datasets. P denotes precision, R the recall, and F1 the Macro F1 score.}
\label{fig:confusion_matrices}
\begin{subfigure}{0.45\columnwidth}
\centering
\includegraphics[width=\linewidth]{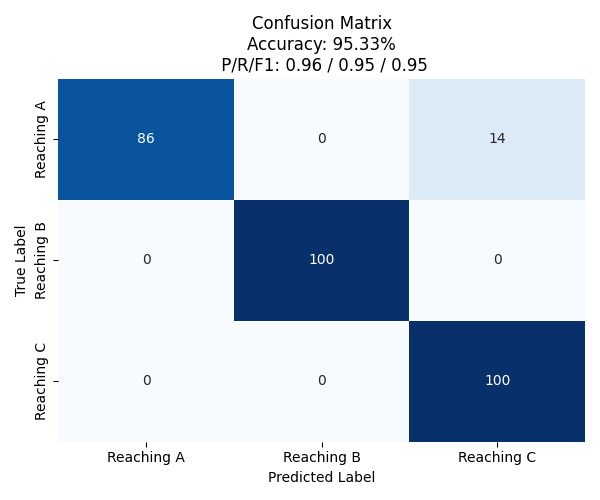}
\caption{$\textsf{I}\to\textsf{A}\textsf{B}\textsf{C}$}
\end{subfigure}
\begin{subfigure}{0.45\columnwidth}
\centering
\includegraphics[width=\linewidth]{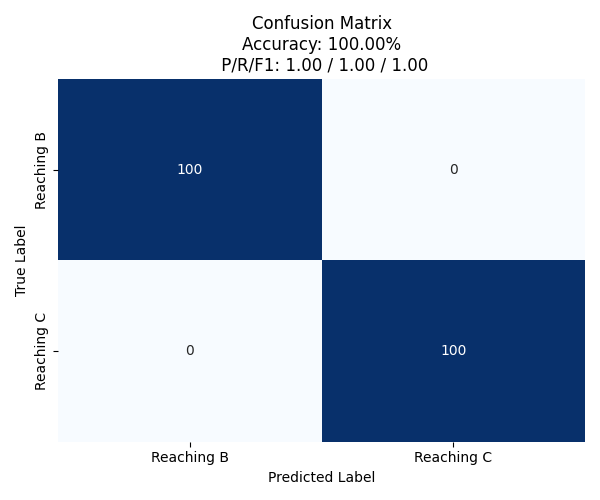}
\caption{$\textsf{A}\to\textsf{BC}$}
\end{subfigure}
\begin{subfigure}{0.45\columnwidth}
\centering
\includegraphics[width=\linewidth]{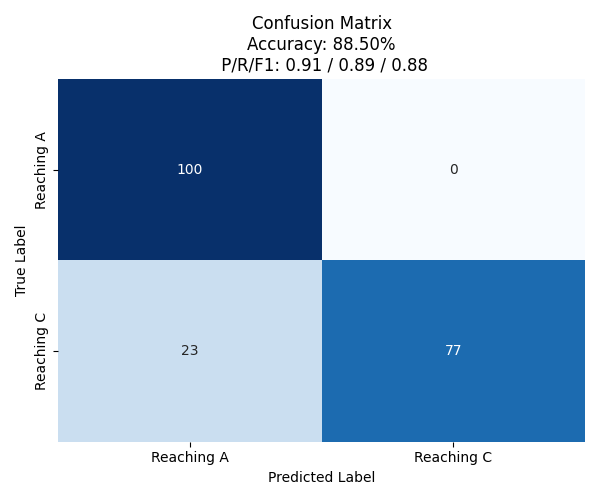}
\caption{$\textsf{B}\to\textsf{AC}$}
\end{subfigure}
\begin{subfigure}{0.45\columnwidth}
\centering
\includegraphics[width=\linewidth]{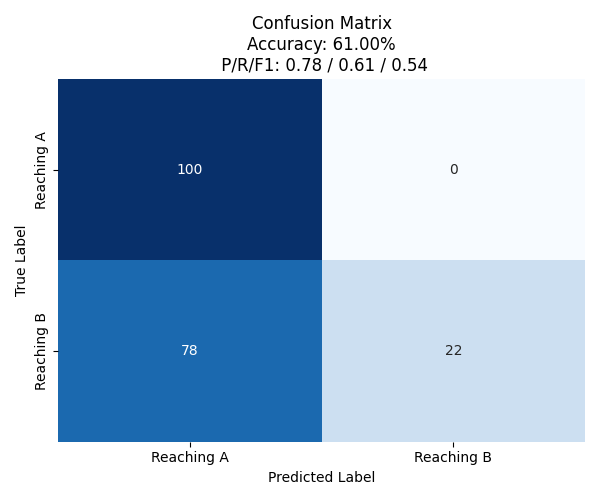}
\caption{$\textsf{C}\to\textsf{AB}$}
\end{subfigure}
\\
\begin{subfigure}{0.45\columnwidth}
\centering
\includegraphics[width=\linewidth]{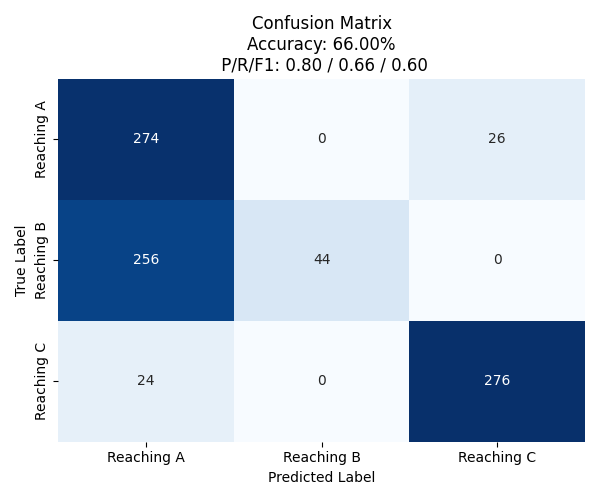}
\caption{$\textsf{I}\to\textsf{A}\textsf{B}\textsf{C}$}
\end{subfigure}
\begin{subfigure}{0.45\columnwidth}
\centering
\includegraphics[width=\linewidth]{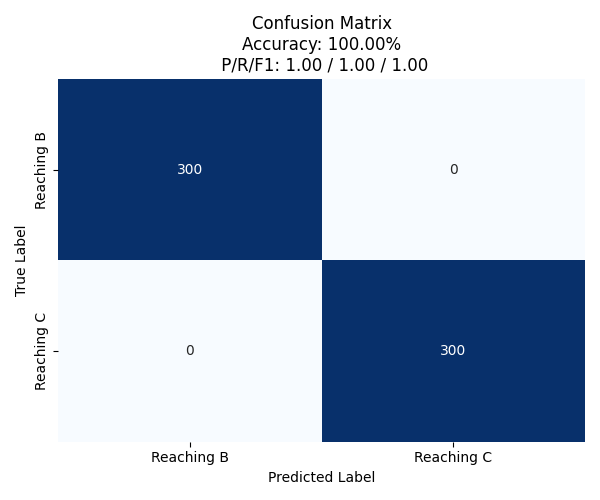}
\caption{$\textsf{A}\to\textsf{BC}$}
\end{subfigure}
\begin{subfigure}{0.45\columnwidth}
\centering
\includegraphics[width=\linewidth]{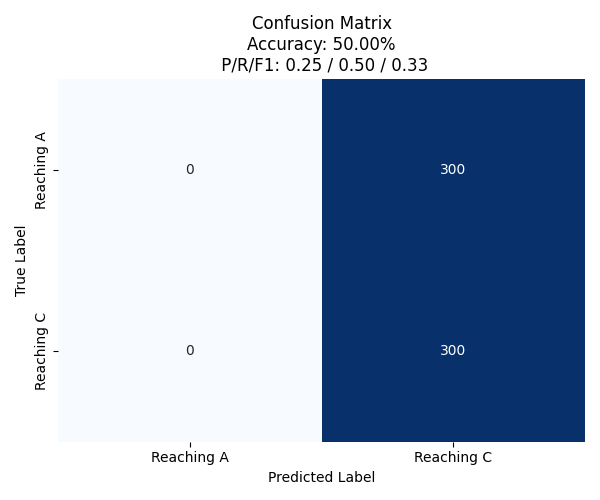}
\caption{$\textsf{B}\to\textsf{AC}$}
\end{subfigure}
\begin{subfigure}{0.45\columnwidth}
\centering
\includegraphics[width=\linewidth]{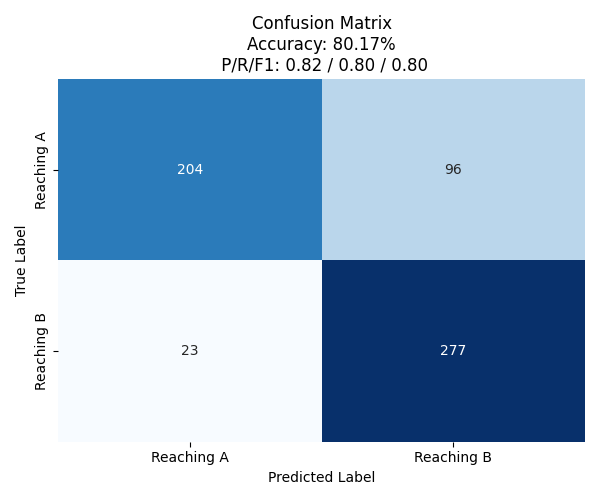}
\caption{$\textsf{C}\to\textsf{AB}$}
\end{subfigure}
\end{figure*}

\noindent
\textit{Results and interpretation}.
In the incremental prefix condition, $\textsf{I}\to\textsf{ABC}$ trained on \textsf{D-0} reaches its maximum accuracy after approximately $170$ samples, whereas the \textsf{D-0-10-30} version requires about $230$ samples.
For $\textsf{A}\to\textsf{BC}$, the \textsf{D-0} classifier reaches peak accuracy after only $30$ samples, while the \textsf{D-0-10-30} version requires about $130$ samples (still before half of the trajectory is observed).
Lower predictive capability is observed for $\textsf{B}\to\textsf{AC}$: for \textsf{D-0}, accuracy saturates only when almost the full trajectory is available ($\approx 250$ out of $251$ samples), while for \textsf{D-0-10-30} the accuracy remains bounded below the noise-free case even with full observation.
For $\textsf{C}\to\textsf{AB}$, the \textsf{D-0} version peaks after $\approx 220$ samples, whereas the \textsf{D-0-10-30} version requires $\approx 240$ samples.
These trends align with the movement prediction sensitivity analysis.
As observation noise increases, more evidence is required for confident semantic disambiguation.
From an HRC perspective, this implies that under degraded sensing, intent prediction may be delayed or uncertain, requiring the robot to adopt conservative assistance and safety policies until confidence stabilises.
In the sliding window condition, the overall pattern is consistent with the incremental prefix setting, but performance is more sensitive to local trajectory features because long-range temporal context is not retained.
Results are summarised in Table~\ref{tab:accuracy} and Figure~\ref{fig:confusion_matrices}.
Beyond overall accuracy, the confusion matrices reveal important class-dependent effects that are better captured by precision and recall. 
In \textsf{D-0}, all classes achieve high recall (above $0.86$), yielding a F1-score of approximately $0.95$.
In contrast, considering \textsf{D-0-10-30}, the degradation is highly asymmetric. 
While $\textsf{A}$ and $\textsf{C}$ maintain recall above $0.9$, the recall of $\textsf{B}$ drops to $0.15$, despite retaining perfect precision. 
The model becomes extremely conservative in predicting $\textsf{B}$.
When $\textsf{B}$ is predicted it is correct, but most true $\textsf{B}$ instances are misclassified.
This deserves a more careful ablation analysis in future work, for example, separate evaluation per noise level, and controlled regularisation sweeps.

\noindent
\textit{Computational cost}.
For each classifier, we compute average and standard deviation prediction times across the test set.
Results are shown in Table~\ref{tab:prediction_times}.
Times are in the microsecond range and show negligible variance, suggesting that discrete belief updates can be performed at high rates and are compatible with tight HRC timing constraints.

\begin{table}[t]
\centering
\caption{Prediction times for action prediction/classification models.}
\label{tab:prediction_times}
\begin{tabular}{ccc|ccc}
\multicolumn{3}{c}{\textsf{D-0}} & \multicolumn{3}{c}{\textsf{D-0-10-30}} \\
\hline
\hline
\textbf{Model} & \textbf{Avg.} [s] & \textbf{Std.} [s] & \textbf{Model} & \textbf{Avg.} [s] & \textbf{Std.} [s] \\ 
\hline
$\textsf{I}\to\textsf{A}\textsf{B}\textsf{C}$ & 0.0007 & 0.0003 & $\textsf{I}\to\textsf{A}\textsf{B}\textsf{C}$ & 0.0006 & 0.0000 \\
$\textsf{A}\to\textsf{BC}$ & 0.0006 & 0.0000 & $\textsf{A}\to\textsf{BC}$ & 0.0006 & 0.0000 \\
$\textsf{B}\to\textsf{AC}$ & 0.0006 & 0.0000 & $\textsf{B}\to\textsf{AC}$ & 0.0006 & 0.0000 \\
$\textsf{C}\to\textsf{AB}$ & 0.0006 & 0.0000 & $\textsf{C}\to\textsf{AB}$ & 0.0006 & 0.0000 \\
\hline
\end{tabular}
\end{table}

\noindent
\textit{An end-to-end example}.
We discuss a simplified example illustrating the qualitative behaviour of \textsf{MA-HERP} as described in Algorithm~\ref{alg:hrf}.
We randomly generate a sequence of $15$ consecutive reaching movements (Table~\ref{tab:predicted_labels}), which can be seen as surrogates of simple pick-and-place behaviours in HRC, where multiple reaches must be chained coherently and predictions must be updated on-line.
Initially, the wrist is at pose \textsf{I}.
At each step, a new target volume is sampled uniformly from \{\textsf{A},\textsf{B},\textsf{C}\}, excluding the previously selected one.
This induces a prior over the active motion label $\ell_k$, a trajectory is generated accordingly, and the paired movement predictor and action classifier conditioned on the same label are activated.

\begin{table}[t]
\centering
\caption{Predicted \textit{versus} actual labels for the sequence of $15$ \texttt{reach} movements.}
\label{tab:predicted_labels}
\begin{tabular}{c|c|c|c}
\multicolumn{1}{c}{\textbf{Movement}} & 
\multicolumn{1}{c}{\textbf{$\alpha(\ell_k)$}} &
\multicolumn{1}{c}{\textbf{Accuracy}} & 
\multicolumn{1}{c}{\textbf{$\ell_k$}} \\
\hline
\hline
$\textsf{I}\to\textsf{C}$ & \texttt{reach-c} & 0.73 & \texttt{reach-c} \\ 
$\textsf{C}\to\textsf{A}$ & \texttt{reach-a} & 0.81 & \texttt{reach-a} \\
$\textsf{A}\to\textsf{B}$ & \texttt{reach-b} & 1.00 & \texttt{reach-b} \\
$\textsf{B}\to\textsf{A}$ & \texttt{reach-a} & 1.00 & \texttt{reach-a} \\
$\textsf{A}\to\textsf{B}$ & \texttt{reach-b} & 1.00 & \texttt{reach-b} \\
$\textsf{B}\to\textsf{C}$ & \texttt{reach-c} & 0.54 & \texttt{reach-c} \\
$\textsf{C}\to\textsf{B}$ & \texttt{reach-a} & 0.42 & \texttt{reach-b} \\
$\textsf{B}\to\textsf{C}$ & \texttt{reach-c} & 0.54 & \texttt{reach-c} \\
$\textsf{C}\to\textsf{A}$ & \texttt{reach-a} & 1.00 & \texttt{reach-a} \\
$\textsf{A}\to\textsf{B}$ & \texttt{reach-b} & 1.00 & \texttt{reach-b} \\
$\textsf{B}\to\textsf{C}$ & \texttt{reach-c} & 0.92 & \texttt{reach-c} \\
$\textsf{C}\to\textsf{A}$ & \texttt{reach-a} & 1.00 & \texttt{reach-a} \\
$\textsf{A}\to\textsf{C}$  & \texttt{reach-c} & 1.00 & \texttt{reach-c} \\
$\textsf{C}\to\textsf{A}$ & \texttt{reach-a} & 0.58 & \texttt{reach-a} \\
$\textsf{A}\to\textsf{B}$ & \texttt{reach-b} & 1.00 & \texttt{reach-b} \\
\hline
\end{tabular}
\end{table}

Assuming that $X_t$ is received on-line in chunks of increasing size (incremental prefix condition), once the first $W=100$ samples are acquired the movement predictor generates $P=50$-step forecasts $\widehat{X}_{t:t+P}$ at each time step.
Each forecasted segment is then provided to the action classifier, which produces the instantaneous belief $\alpha_t(\ell)$ and can, in a full \textsf{MA-HERP} setting, be regularised by admissible temporal relations via $\psi_{\mathcal{R}}$ and hierarchical consistency via $\psi_{\mathcal{C}}$.
When the end of the trajectory is reached, majority voting yields a final discrete label, and a new target is selected for the next movement.
This illustrates the intended HRC usage pattern: discrete intent estimates are continuously updated during motion, enabling the robot to prepare early responses (e.g., safe postures or compatible assistance) without waiting for completion.

\begin{figure*}[t!]
\centering
\caption{
Configuration spaces of the first three \texttt{reach} movements in the sequence, namely $\textsf{I}\to\textsf{C}$, $\textsf{C}\to\textsf{A}$, and $\textsf{A}\to\textsf{B}$, with \textsc{before} relations between consecutive movements.
Black: generated motion. Red: predicted trajectories.}
\label{fig:consecutive_actions}
\begin{subfigure}{0.64\columnwidth}
    \centering
    \includegraphics[width=\linewidth]{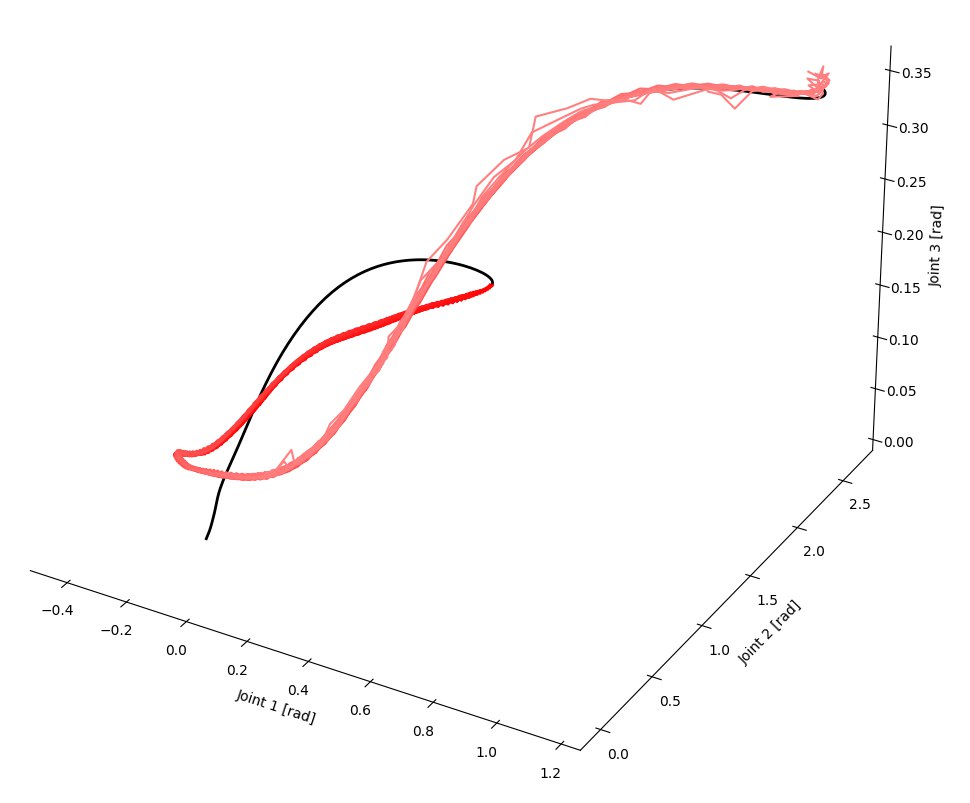}
    \caption{$\textsf{I}\to\textsf{C}$}
\end{subfigure}
\begin{subfigure}{0.64\columnwidth}
    \centering
    \includegraphics[width=\linewidth]{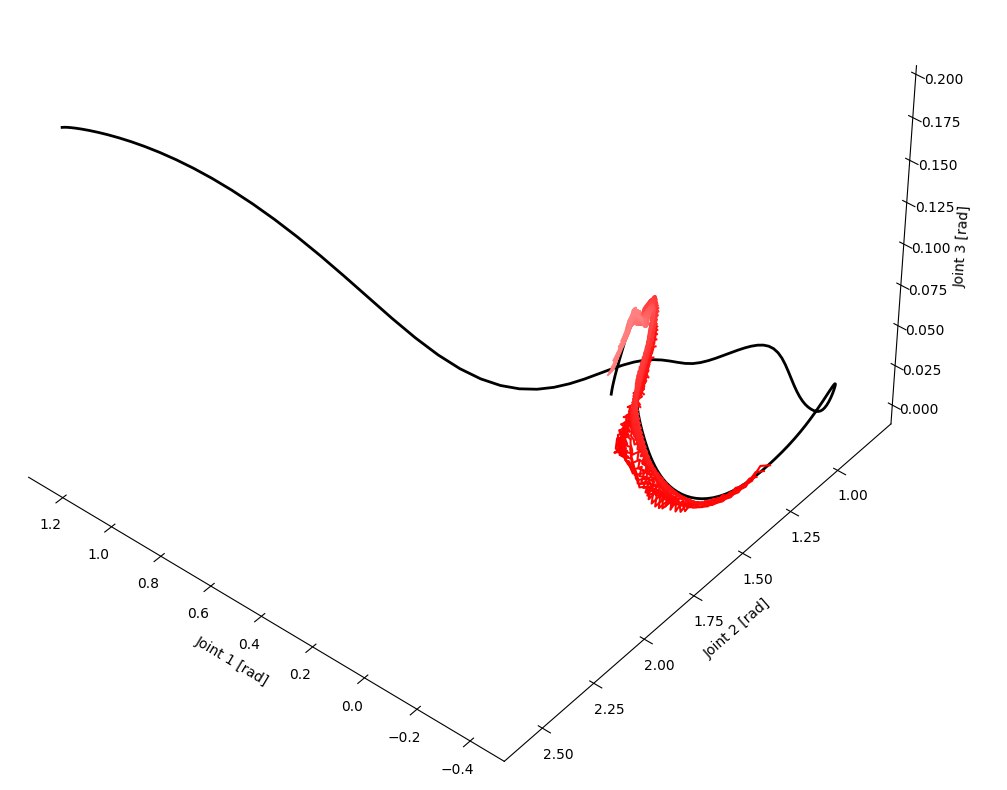}
    \caption{$\textsf{C}\to\textsf{A}$}
\end{subfigure}
\begin{subfigure}{0.64\columnwidth}
    \centering
    \includegraphics[width=\linewidth]{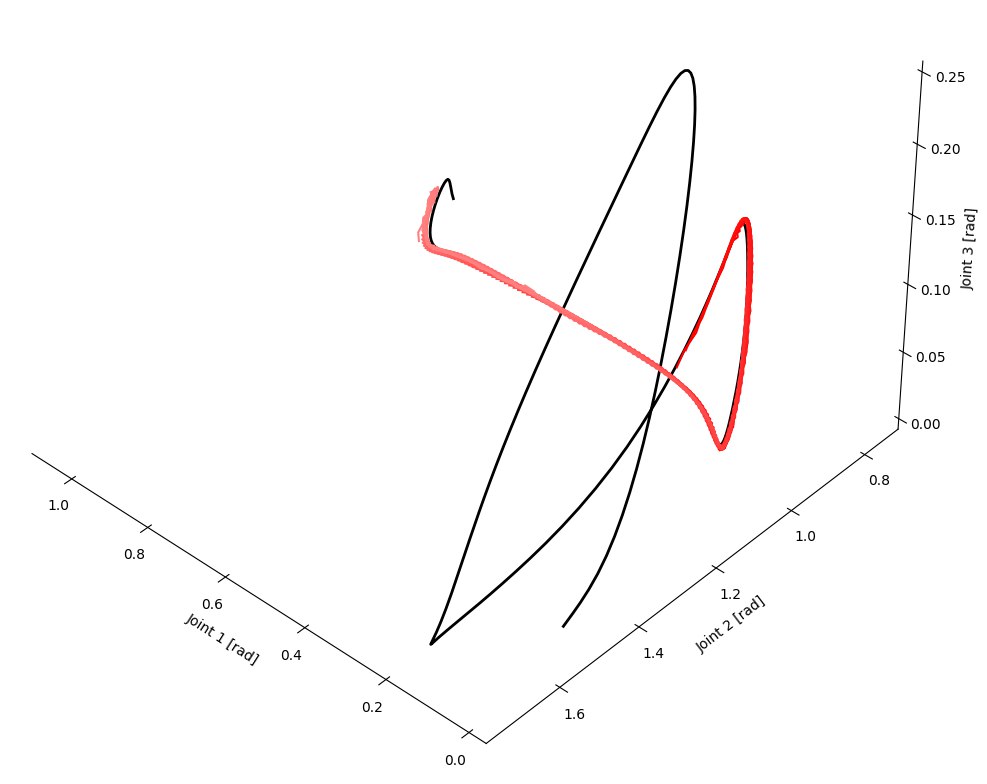}
    \caption{$\textsf{A}\to\textsf{B}$}
\end{subfigure}
\end{figure*}

\section{Conclusions}
\label{sec:conclusions}

In this paper we present \textsf{MA-HERP}, a hierarchical and recursive framework for the \emph{joint} estimation and prediction of human movements and discrete actions, motivated by the fluency requirements of human--robot collaboration. 
\textsf{MA-HERP} integrates 
(i) an interval-based hierarchical representation of actions and temporal relations, 
(ii) a unified factorized model coupling continuous dynamics and discrete labels, and 
(iii) a recursive inference scheme that alternates prediction and evidence assimilation while enforcing structural constraints.

Our evaluation supports three preliminary conclusions.
First, label-conditioned motion prediction can be highly accurate in the noise-free regime and degrades predictably under observation perturbations, which motivates explicit uncertainty handling when embedded in real HRC pipelines.
Second, intent inference can converge before motion completion for several initial conditions, which is precisely the regime in which predictive modules become valuable for \emph{fluency}.
Early commitment enables proactive assistance and timely coordination, while delayed or unstable beliefs call for conservative robot behaviours.
Third, the computational costs of both movement forecasting and discrete belief updates are compatible with on-line operation, suggesting that \textsf{MA-HERP} can be integrated in real-time HRC loops where tight timing and safety constraints make reactive-only strategies insufficient.

Several limitations remain.
First, the current biomechanical setup is intentionally simplified and should be extended to richer kinematics, contact-rich motions, and multi-effector behaviours.
Second, a systematic real-time assessment in closed-loop interaction is required to validate robustness under realistic sensing, latency, and partial observability.
Third, the trade-off between history length and prediction horizon ($W$ and $P$) should be analysed more formally, together with uncertainty calibration.
Finally, the interaction between motion forecasting and iterative action re-labelling should be characterised to avoid oscillations and to support conservative decision-making when beliefs are unstable.

\bibliographystyle{IEEEtran}
\bibliography{AIM2026/AIM2026}

\end{document}